\documentclass[twocolumn]{svjour3}          
\smartqed  
\usepackage{graphicx}
\usepackage[numbers]{natbib}
\usepackage{times}
\usepackage{epsfig}
\usepackage{amsmath}
\usepackage{amssymb}
\usepackage{lineno}
\usepackage[dvipsnames]{xcolor}

\usepackage[pagebackref=true,breaklinks=true,letterpaper=true,colorlinks,bookmarks=false,citecolor=blue,linkcolor=blue]{hyperref}

\usepackage{booktabs}
\usepackage{amsmath}
\usepackage{amssymb}
\usepackage{verbatim}
\usepackage{multirow, makecell}
\usepackage{lineno}
\usepackage{enumerate}
\usepackage{amsfonts}
\usepackage{gensymb}
\usepackage{bm}
\usepackage{bbding}
\usepackage{comment}
\usepackage{color}
\usepackage{diagbox}
\usepackage{tabularx}
\usepackage{rotating}
\usepackage{listings}
\usepackage{pifont}
\newcolumntype{L}[1]{>{\raggedright\let\newline\\\arraybackslash\hspace{0pt}}m{#1}}
\newcolumntype{C}[1]{>{\centering\arraybackslash}m{#1}}
\newcolumntype{R}[1]{>{\raggedleft\let\newline\\\arraybackslash\hspace{0pt}}m{#1}}

\newlength\savewidth

\usepackage{colortbl}
\definecolor{Gray}{gray}{0.9}
\definecolor{GrayT}{gray}{0.4}

\usepackage{subcaption}
\usepackage[labelfont=bf]{caption}

\definecolor{light-blue}{RGB}{186,175,255}
\definecolor{light-red}{RGB}{255,137,165}
\usepackage{arydshln} 
\usepackage{booktabs}
\usepackage{multirow}
\usepackage{amsmath}
\usepackage{etoolbox}
\usepackage{bm}
\usepackage{mathtools}
\usepackage{algorithm}
\usepackage{algpseudocode}
\usepackage{xspace}
\usepackage{xparse}
\usepackage{multirow, makecell}
\usepackage{colortbl}  
\usepackage{xcolor}
\usepackage{array}
\definecolor{Gray}{gray}{0.94}
\definecolor{liGray}{gray}{0.5}
\definecolor{LightCyan}{rgb}{0.88,1,1}
\usepackage{mathrsfs}

\newcommand{\f}{\texttt{f}}
\newcommand{\bo}{\texttt{b}}
\newcommand{\h}{\texttt{h}}
\newcommand{\data}{\texttt{HoCo}}

\begin{document}
\begin{sloppypar}

\title{Beyond Talking -- Generating Holistic 3D Human Dyadic Motion for Communication}

\author{Mingze Sun 
\and 
Chao Xu 
\and
Xinyu Jiang 
\and
Yang Liu       
\and
Baigui Sun 
\and 
Ruqi Huang
}

\institute{
Mingze Sun \and Xinyu Jiang 
\at
Tsinghua Shenzhen International Graduate School, China \\ 
\email {\{smz22, xy-jiang23\}@tsinghua.edu.cn} \\ 
\and
Chao Xu \and Yang Liu \and Baigui Sun \at Alibaba Group  \\ 
\email {\{xc264362, ly261666, baigui.sbg\}@alibaba-inc.com} \\ 
\and
Ruqi Huang {*corresponding author}
\at
Tsinghua Shenzhen International Graduate School, China \\ 
\email {\{ruqihuang\}@sz.tsinghua.edu.cn} 
}

\date{Received: date / Accepted: date}



\maketitle
\begin{abstract}
In this paper, we introduce an innovative task focused on human communication, aiming to generate 3D holistic human motions for both speakers and listeners. Central to our approach is the incorporation of factorization to decouple audio features and the combination of textual semantic information, thereby facilitating the creation of more realistic and coordinated movements. We separately train VQ-VAEs with respect to the holistic motions of both speaker and listener. We consider the real-time mutual influence between the speaker and the listener and propose a novel chain-like transformer-based auto-regressive model specifically designed to characterize real-world communication scenarios effectively which can generate the motions of both the speaker and the listener simultaneously. These designs ensure that the results we generate are both coordinated and diverse. Our approach demonstrates state-of-the-art performance on two benchmark datasets. Furthermore, we introduce the \texttt{HoCo} holistic communication dataset, which is a valuable resource for future research. Our \texttt{HoCo} dataset and code will be released for research purposes upon acceptance.
\end{abstract}

\section{Introduction}
\label{sec:intro}

\begin{figure}[!b]
  \begin{center}
    \includegraphics[width=0.5\textwidth]{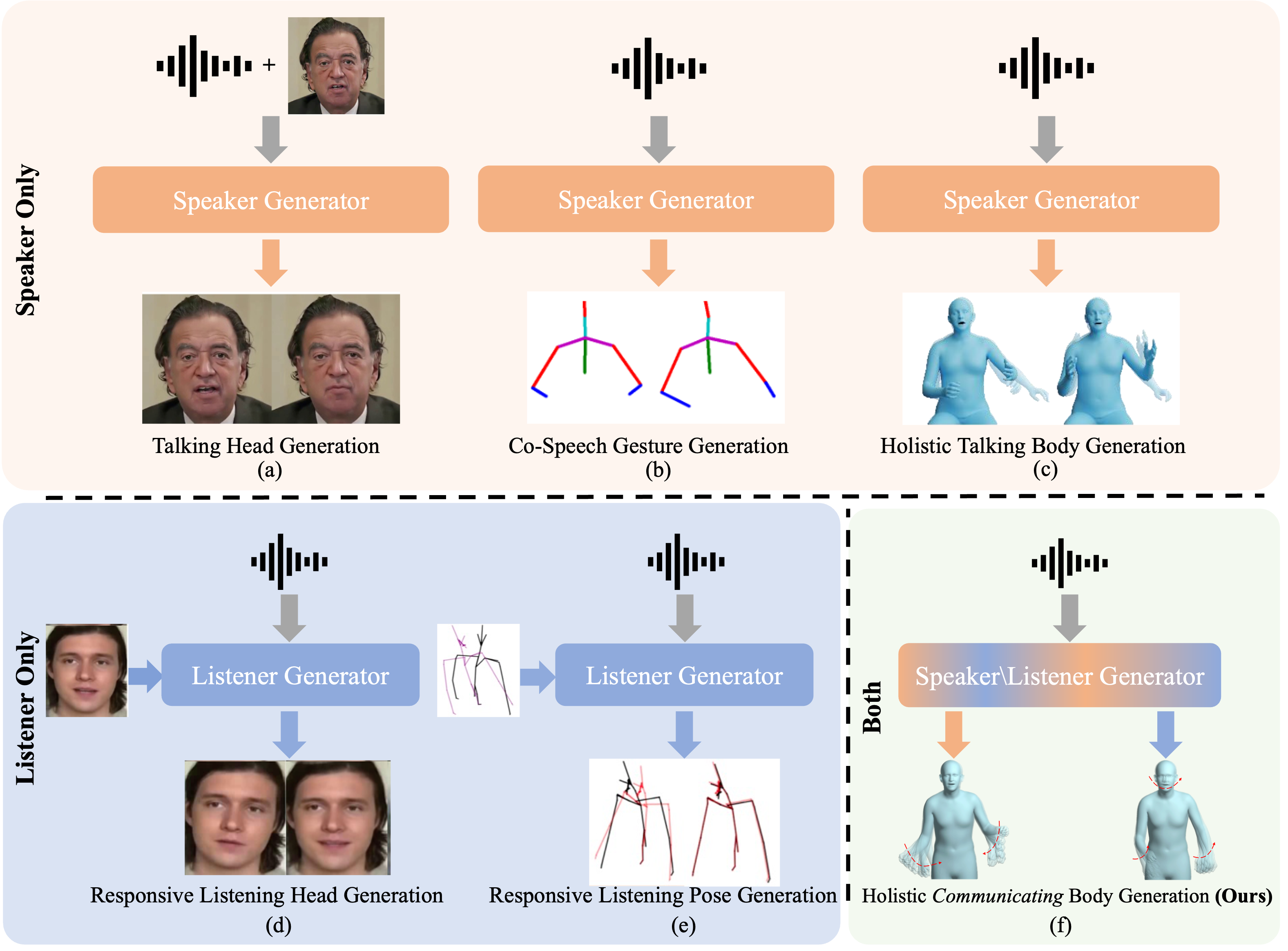}
  \end{center}
\caption{Illustrations of two related tasks and our proposed holistic communicating body generation. Top row: Generation of the head~\cite{zhang2023sadtalker}, gesture~\cite{zhi2023livelyspeaker}, or holistic body~\cite{yi2022generating} for the speaker from speech; Bottom left: Responsive listening head synthesizes videos in responding to the speaker video stream~\cite{zhou2022responsive} and Responsive listening pose generation based on predicted pose history~\cite{ahuja2019react}; Bottom right: Our holistic communicating body generation from speech. We generate 3D holistic motions for both speakers and listeners simultaneously.}
\label{fig:task}
\end{figure}

\begin{figure*}[t!]
  \begin{center}
    \includegraphics[width=\textwidth]{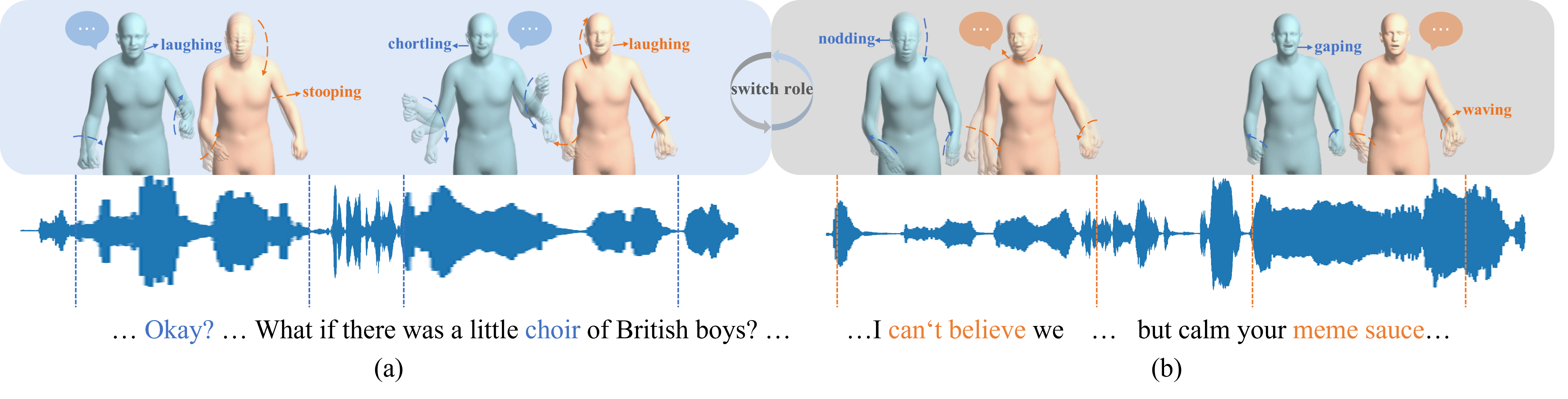}
  \end{center}
\caption{Holistic Communicating Body Generation Example. Given a talk between two participants, our method can generate coordinated and diverse communication. In (a) the speaker is on the left, and the listener is on the right. The listener laughs in response to the speaker's joke, accompanied by changes in body posture. In (b), the roles are switched. The speaker is on the right, and the listener is on the left. As the speaker narrates an unusual event, the listener expresses surprise with raised gestures and facial expressions in sync with the speaker.}
\label{fig:teaser}
\end{figure*}
Boosted by large-scale video data of human talking, recent approaches have made significant progress in the task of speech-to-motion, namely, generating non-verbal signals (\emph{e.g.}, facial expressions or body movements) within a talk, such as human facial expressions, body poses, and hand gestures, from verbal cues (\emph{e.g.}, audio clips or transcript). 
Such progress can help artificial intelligence agents understand human behavior within a speech, and therefore contributes to practical applications in healthcare, virtual reality (VR), and human-robot interaction (HRI), to name a few.

However, the prior works all concentrate on playing a \emph{single role}, either speaker or listener, in a talk. 
For the former, there have been works on the alignment of facial and lip movements with audio~\cite{ye2023geneface, zhang2023sadtalker, guo2021ad, xu2023high} as shown in Fig.~\ref{fig:task}(a), realistic gesture generation~\cite{ao2022rhythmic, zhi2023livelyspeaker, ahuja2023continual} in Fig.~\ref{fig:task}(b), as well as the synthesis of holistic 3D human motion for speakers~\cite{habibie2021learning, yi2022generating} in Fig.~\ref{fig:task}(c); For the latter, the recent advances have been primarily devoted to generating facial expression reactions~\cite{zhou2022responsive, song2023emotional, liu2023mfr} in Fig.~\ref{fig:task}(d) and the body poses of the listeners~\cite{tuyen2022agree, tuyen2023takes} in Fig.~\ref{fig:task}(e). However, these works cannot directly generate holistic 3D mesh motions for \emph{both speaker and listener simultaneously} based on audio alone.  

On the other hand, it has long been recognized that \textbf{communication} is of significance in social interaction between humans. 
Early research~\cite{AO, Birdwhistell} even argues that over $80\%$ of human communication is encoded in facial expressions and body movements. 
In fact, participants of a conversation can simultaneously convey and perceive information throughout, no matter whether he/she is talking or not. 
For instance, when a speaker says something confusing, a listener may have a furrowed brow (facial expression) and/or rub head with fingers (body movement). The speaker can then adjust his/her expression by noticing the listener's reaction. 
Since the prior works, all generate non-verbal signals of a fixed role, it is evident that none can take into consideration the instantaneous mutual influences among participants. 
As a consequence, the agent trained in this single-role manner lacks the potential of \emph{real-time interacting} with other agents, let alone humans in real life. 

In this paper, we present a significant advancement towards the generation of holistic motions for \emph{both} speaker and listener.
As shown in Fig.~\ref{fig:task}, our approach significantly differs from the single-role based generative methods. 
More specifically, given a talk between two participants, we take consideration of the interaction in between and aim to simultaneously generate holistic 3D mesh sequences including facial expressions, body poses, and hand gestures, for both speaker and listener. 
Fig.~\ref{fig:teaser} shows an example that our method is capable of generating diverse and coordinated communication during a conversation between two participants switching roles. 

To achieve this goal, we first propose \texttt{HoCo} dataset, which provides \texttt{Ho}listic description of \texttt{Co}mmunication within talks, including verbal (\emph{e.g.}, audio and transcript) and non-verbal signals (\emph{e.g.}, holistic body language) of both speakers and listeners.  
More concretely, we collect $22,913$ pieces of video clips, whose total duration is $45$ hours. 
We further make our effort to provide comprehensive annotations for both speakers and listeners within the video clips including audio, text aligned with the speech, and the pseudo-labels for SMPL-X. 

On the front of modeling, our novel task requires effective modeling of not only each role but also their mutual influence. 
For the former, we propose to enhance our model by introducing proper factorization of the verbal input, which in turn reduces the complexity of the latent space and allows for more fine-grained control over generation. The existing methods~\cite{kucherenko2020gesticulator, bhattacharya2021speech2affectivegestures} encode information from both audio and corresponding text. 
However, the features have not been decoupled based on the influence of facial expressions and body motions. For instance, the shape of the speaker's mouth should be related to the content and style of the input audio, while body motions should consider more of the style of the audio and the semantic information of the corresponding text.
In contrast to prior methods, we resort to the recent advance in pre-training audio feature factorization~\cite{li2022styletts}, which encodes input audio into features consisting of three components in correspondence to attributes including energy, pitch, and style. 
Beyond the audio rhythm and style, since human motions are related to the semantic information of the text and our dataset provides accompanied texts with respect to the verbal input, we further enrich the encoding by appending the respective features extracted by the pre-trained NLP model of~\cite{devlin2018bert}. 

For the latter, while it seems plausible to combine the existing works on speakers and that on listeners to generate non-verbal interaction, we argue that this naive approach can not fully exploit the interaction presented in the data and leads to sub-optimal results in the end. That is because a) works on speaker~\cite{yi2022generating, yoon2020speech, liu2022learning, guo2021ad} typically ignores listener; b) works on listener~\cite{zhou2022responsive, song2023emotional, liu2023mfr} most require the ground-truth annotations of speakers during inference, which is inconvenient but also favors speakers' impact over listeners much more than the opposite. 

In contrast, our method considers the real-time mutual influence between the speaker and the listener. We first separately train VQ-VAEs~\cite{van2017neural} with respect to the body language of both speaker and listener. 
Subsequently, we propose an auto-regressive transformer model with a tailored chain-like structure design. 
More concretely, we condition the generation of the body language of the speaker at frame $T+1$ on the factorized features and the generated frames from frame $1$ to frame $T$ of both speaker and listener. 
The counterpart for the listener at frame $T+1$ is similarly conditioned but with an extra input of the speaker at frame $T+1$, which reflects the slight advantage of speakers in leading communication. 
\begin{table*}[!t]
\caption{Comparison of different speech-to-motion datasets. The horizontal line dividers represent, respectively: non-interactive lab scenes, non-interactive wild scenes, interactive scenes and our provided \texttt{HoCo} dataset.}\label{table:data}
\centering
\setlength{\tabcolsep}{16pt}
\resizebox{\textwidth}{24mm}{
\begin{tabular}{lllllllll}
\hline
\rowcolor[HTML]{FFFFFF} 
Dataset         & Environment & Interaction & Holistic Body Connection & Body        & Head      & Hand        & Annotation   & Length \\ \hline
\rowcolor[HTML]{FFFFFF} 
Multiface~\cite{wuu2022multiface}       & Lab         & \ding{55}           & \ding{55}                        & \ding{55}           & 3D mesh   & \ding{55}           & multi-camera & /      \\
\rowcolor[HTML]{FFFFFF} 
VOCASET~\cite{cudeiro2019capture}         & Lab         & \ding{55}           & \ding{55}                        & \ding{55}           & 3D mesh   & \ding{55}           & 4D-scan      & /      \\
\rowcolor[HTML]{FFFFFF} 
Takeuchi et.al~\cite{takeuchi2017creating}  & Lab         & \ding{55}           & \ding{55}                        & 3D keypoint & \ding{55}         & \ding{55}           & MoCap        & 5h     \\
\rowcolor[HTML]{FFFFFF} 
Trinity~\cite{ferstl2018investigating}         & Lab         & \ding{55}           & \ding{55}                        & 3D keypoint & \ding{55}         & \ding{55}           & MoCap        & 4h     \\ \hline
\rowcolor[HTML]{FFFFFF} 
Yoon et.al~\cite{yoon2019robots, yoon2020speech}      & Wild        & \ding{55}           & \ding{55}                        & 3D keypoint & \ding{55}         & \ding{55}           & p-GT         & 52h    \\
\rowcolor[HTML]{FFFFFF} 
Speech2Gesture~\cite{ginosar2019learning}  & Wild        & \ding{55}           & \ding{55}                        & 2D keypoint & \ding{55}         & 2D keypoint & p-GT         & 144h   \\
\rowcolor[HTML]{FFFFFF} 
Habibie et.al~\cite{habibie2021learning}     & Wild        & \ding{55}           & \ding{55}                        & 3D keypoint & 3D mesh   & 3D keypoint & p-GT         & 33h    \\
\rowcolor[HTML]{FFFFFF} 
SHOW~\cite{yi2022generating}            & Wild        & \ding{55}           & \Checkmark                        & 3D mesh     & 3D mesh   & 3D mesh     & p-GT         & 27h    \\ \hline
\rowcolor[HTML]{FFFFFF}
UDIVA~\cite{palmero2021context}                   & Lab         & \Checkmark                    & \ding{55}                        & RGB video   & RGB video  & \ding{55}            & multi-camera  & 90.5h         \\
\rowcolor[HTML]{FFFFFF} 
Talking With Hands~\cite{lee2019talking}      & Lab         & \Checkmark                    & \ding{55}                        & 3D keypoint & \ding{55}          & 3D keypoint  & multi-camera  & 50h           \\
\rowcolor[HTML]{FFFFFF} 
JESTKOD~\cite{bozkurt2017jestkod}                 & Lab         & \Checkmark                    & \ding{55}                        & 3D keypoint & \ding{55}          & \ding{55}            & multi-camera  & 4.3h          \\
\rowcolor[HTML]{FFFFFF} 
LISI-HHI~\cite{tuyen2023multimodal}                & Lab         & \Checkmark                    & \ding{55}                        & 3D keypoint & \ding{55}          & \ding{55}            & multi-camera  & 8.3h          \\
\rowcolor[HTML]{FFFFFF} 
ViCo~\cite{zhou2022responsive}            & Wild        & \Checkmark           & \ding{55}                        & \ding{55}           & RGB video & \ding{55}           & p-GT         & 1.5h   \\
\rowcolor[HTML]{FFFFFF} 
Learning2Listen~\cite{ng2022learning} & Wild        & \Checkmark           & \ding{55}                        & \ding{55}           & 3D mesh   & \ding{55}           & p-GT         & 72h    \\ \hline
\rowcolor[HTML]{E7E6E6} 
Ours            & Wild        & \Checkmark           & \Checkmark                        & 3D mesh     & 3D mesh   & 3D mesh     & p-GT         & 45h    \\ \hline
\end{tabular}}
\vspace{-0.5em}
\end{table*}

We demonstrate both quantitative and qualitative experimental comparisons with respect to the prior state-of-the-art. Remarkably, our model generates more coordinated and diverse single-role body language, and we achieve a $27.6$\% improvement in Frechet Gesture Distance (\textbf{FGD}) and a $46.2$\% improvement in \textbf{Variation} compared to~\cite{yi2022generating}. 
We also significantly outperform the baselines in the more challenging communication generation between two participants, which is benchmarked on our proposed dataset. 

To summarize, our contributions are as follows: 
\begin{itemize}
    \item We present \texttt{HoCo} dataset, which comprises high-definition RGB videos of communication within human interaction. The dataset spans 45 hours and includes 22,913 video clips, accompanied by multi-modal corresponding information. We have also generated pseudo-labels for SMPL-X in generating full-body movements for both speakers and listeners.
    \item We design a new audio decoupling method that incorporates text information as conditions for generating both speaker and listener motions. The decoupled features include content, style, and semantic information, which correspond to the control of generating expressions and body motions. This results in motions that exhibit stronger consistency with the audio phonologically and semantically.
    \item We devise a framework for the chained generation of speaker and listener real-time interactions, allowing for the simultaneous generation of full-body movements for both speakers and listeners based on a given raw audio during inference. The generated speaker and listener motions exhibit a more coordinated and cohesive interaction.
\end{itemize}

\section{Related work}
\label{sec:related}
\subsection{Audio-based Motion Generation}
\noindent\textbf{Speaker-Centric Generation from Speech} has attracted considerable research interest in recent years. 
The initial efforts have been made towards talking head generation within 2D~\cite{wang2021one, zhang2023sadtalker, doukas2021headgan} and 3D~\cite{cudeiro2019capture, richard2021meshtalk, fan2022faceformer} domain. 
Another line of work addresses the generation of body poses of speakers based on audio, which can be divided into rule-based~\cite{kipp2005gesture, kopp2006towards} and learning-based methods~\cite{huang2014learning, levine2010gesture, sargin2008analysis}, of which we focus on the latter. 
2D or 3D skeletons have been popular representations for body pose generation~\cite{yoon2020speech, zhi2023livelyspeaker, ao2022rhythmic, zhu2023taming}, while being efficient, such representations can only encode body movements in a coarse and abstract manner.  
To this end, full-body motion generalization consisting of body movements and hand gestures~\cite{habibie2021learning, yi2022generating} have been proposed. 
Overall, it is evident that a more and more comprehensive understanding of non-verbal signals has been achieved along this line of work. 
Indeed, our holistic description of non-verbal communication greatly benefits from the recent SOTA~\cite{yi2022generating}. 
The specific task differences can be referenced from the upper part of Fig.~\ref{fig:task}.
Nevertheless, these approaches completely ignore the perspective of listeners, which plays an equally important role in communication. 

\noindent\textbf{Listener-Centric Generation from Speech} primarily focuses on the non-verbal reactions of listeners with respect to speakers' output. 
There has been some progress in recent years on the work concerning the facial reactions of listeners.
Their task can be summarized as shown in Fig.~\ref{fig:task}(d).
Data-driven generation of listeners has been first introduced by~\cite{gillies2008responsive}, which is purely conditioned on the input audio clips. 
More recent breakthrough~\cite{zhou2022responsive} leverages deep neural networks and considers the generation of listener's faces conditioned on both audio and speaker's faces. 
Follow-up works consider more general encoding including the head motion of listeners, and have further improved the performance by utilizing advances on generative models, such as VQ-VAE model~\cite{ng2022learning} and diffusion model~\cite{liu2023mfr}. 
Song et al.~\cite{song2023emotional} also introduce extra emotion information to generate realistic listener head motion. The recent REACT challenge~\cite{song2023react2023} considers online multiple-listener facial reaction generation based on the speaker’s real-time behavior.
Similar to the aforementioned works on speakers, the current efforts in listener generation are also biased. 
Moreover, they all focus on generating facial expressions, which fail to capture the full-body movements. 
A series of works also focus on interlocutor-aware listener body reactions, as shown in Fig.~\ref{fig:task}(e).
Early effort along this line focuses on predicting the listener's response based on the speaker's behavior~\cite{joo2019towards, jonell2020let}. 
Based on these, ~\cite{ahuja2019react} further adds the previous listener's pose to predict the listener's next pose. 
More recently, several works utilize popular generative models like GANs to generate body poses of listeners~\cite{tuyen2022agree, tuyen2023takes}. 
The GENEA challenge~\cite{kucherenko2023genea} considers interlocutor-aware holistic motions, and the DYAD challenge~\cite{palmero2022chalearn} considers behavior forecasting for the upper body, faces, and head of the two interlocutors simultaneously. 
However, these works all use skeletons to represent 3D poses, which lack detailed expression compared to meshes and cannot be directly used for rendering. 
Additionally, previous works only consider generating listener behavior and ignore the mutual influence between the speaker and the listener. 

We advocate considering a new task, as shown in Fig.~\ref{fig:task}(f), that generates 3D holistic motions for \emph{both speakers and listeners simultaneously} based on the input audio while considering the \emph{mutual influence between the speaker and the listener}.

\subsection{Speech-to-Motion Dataset Construction} 

\noindent\textbf{Speech-to-Motion Datasets} are the cornerstone of the generation task. 
We follow the above clustering criterion to categorize them into speaker-based and listener-based. 
Speech-to-motion based on speaker has been extensively studied, resulting in an abundant variety of datasets in this field, which can be further classified into two types according to the main target: facial expressions and body movements. 
There exists a set of large-scale datasets for talking heads, including 2D datasets ~\cite{nagrani2017voxceleb, lu2021live, guo2021ad} and a 3D one~\cite{cudeiro2019capture}. 
Regarding body movements, related datasets have been proposed in~\cite{yoon2019robots, yoon2020speech, ginosar2019learning}. Among them,~\cite{ginosar2019learning} contains 1766 videos sourced from online TED speech videos and gives 3D poses as pseudo-GroundTruth (p-GT). 
Noting that the aforementioned datasets only consider facial expressions or body movements during speech, without taking into account full-body actions, Habibie et al.~\cite{habibie2021learning} introduces a dataset comprising 33 hours of information involving the holistic movements of speakers. 
Building upon this, Yi et al.~\cite{yi2022generating} performed data cleaning and provided the pseudo labels with respect to the well-known SMPL-X model~\cite{SMPL-X:2019}. 

On the other hand, the construction of datasets focusing on listeners is relatively lagged. 
The ViCo dataset~\cite{zhou2022responsive} is the first one to contain rich samples of different listener identities. 
Another concurrent dataset, Learning2Listen~\cite{ng2022learning}, consists of videos of a total duration of 72 hours, collected in the wild, which comes from YouTube with six identities. 

There are relatively few open-source datasets available for human-human interaction.~\cite{lee2019talking} presents Talking With Hands, which is a dataset of two-person face-to-face spontaneous conversations but fails to capture features beyond body and fingers.~\cite{palmero2021context} introduces UDIVA, a non-acted dataset of face-to-face dyadic interactions, where interlocutors perform competitive and collaborative tasks with different behavior elicitation and cognitive workloads. 
However, this dataset only focuses on the upper body movements of individuals. 
Some datasets focusing on gesture generation in specified dyadic interactions are also available. For instance, the JESTKOD~\cite{bozkurt2017jestkod} dataset focuses on agreement and disagreement scenarios; LISI-HHI dataset~\cite{tuyen2023multimodal} focuses on different interaction scenarios from wayfinding to tangram games. These datasets are all conducted in controlled experimental environments, which significantly limit the diversity of facial expressions and movements as they are tailored to specific tasks. 
To this end, we seek \emph{in-the-wild RGB videos} that also include clear facial, hand, and body movements, allowing for regression to corresponding meshes. 
We, therefore, introduce a large-scale in-the-wild interactive communication dataset, \texttt{HoCo}, which comprises video clips, corresponding audio, text, reconstructed SMPL-X p-GT, and annotations indicating the speaker's position and the listener's emotions. 
For specific dataset comparisons, please refer to Tab.~\ref{table:data}.

\noindent\textbf{Parametric Models }play a role in our dataset construction. 
We benefit from the fact that, under such models, human (or human parts) shapes can be represented by finite-dimension latent codes, giving rise to an efficient representation in both optimizations and also a strong constraint on the plausibility of the generation. 
In the task of talking head synthesis, 3D Morphable Models (3DMM)~\cite{paysan20093d, blanz2023morphable} is widely used to assist in learning accurate mouth shapes and facial expressions~\cite{suwajanakorn2017synthesizing, liu2022semantic, zhang2023sadtalker}. SMPL~\cite{SMPL:2015} is a widely used 3D body model that represents human bodies in a deformable and articulated manner. 
SMPL-X~\cite{SMPL-X:2019} is an important extension of SMPL, which allows for extra encoding and control over hands and face. 
In order to achieve a holistic generation of non-verbal communication, we utilize SMPL-X in our dataset construction, as well as model formulation. 
In particular, we adapt the state-of-the-art method for SMPL-X parameter regression PyMAF-X~\cite{pymafx2023} to compute p-GT with respect to SMPL-X parametric model in our \texttt{HoCo} dataset. 

\section{Methodology}
\begin{figure*}[!ht]
  \begin{center}
    \includegraphics[width=\textwidth]{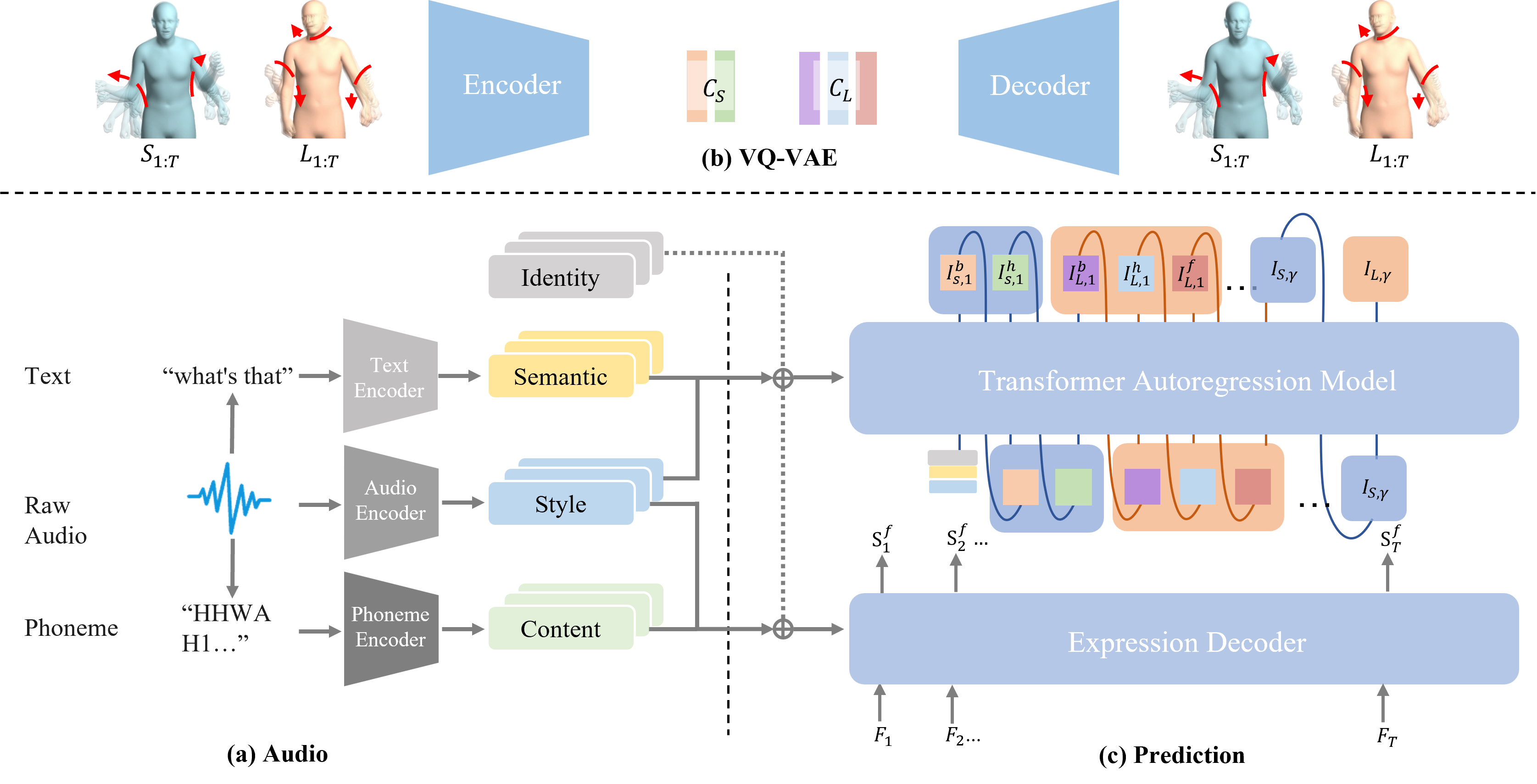}
  \end{center}
\caption{Overview of the proposed framework for holistic communicating generation (a) audio feature extraction (b)VQ-VAE model for the generation of speaker and listener motion (c) Transformer-based autoregression model for simultaneously generating the motion of both speaker and listener in a chain-like manner.}
\label{fig:pipeline}
\end{figure*}

Fig.~\ref{fig:pipeline} shows the schematic illustration of our whole pipeline. 
In the following, we first introduce in Sec.~\ref{sec:pre} how the verbal input is encoded by a mixture of hand-craft feature extractor and pre-trained deep neural networks (Fig.~\ref{fig:pipeline}(a)). Then we present details on the VQ-VAE module for generating the human motion of both speaker and listener in Sec.~\ref{sec:vq}.  (Fig.~\ref{fig:pipeline}(b)). Finally, we propose our auto-regressive transformer for in Sec.~\ref{sec:autoreg} (Fig.~\ref{fig:pipeline}(c)). 

\subsection{Feature Extraction}\label{sec:pre}
The input of our pipeline is assumed to be audio clips as well as the accompanied text. In particular, we follow the same procedure in~\cite{ao2022rhythmic} to align the text with audio clips in the SHOW dataset. 
We assume to be given a audio clip of $T$ frames $A = [a_1, a_2, \cdots, a_T]\in \mathbb{R}^{1\times T}$. The extracted and aligned text is denoted by $W = [w_1, w_2, \cdots, w_T]\in \mathbb{R}^{1\times T}$. 

\noindent\textbf{Identity Encoding: }
Following the previous works~\cite{yi2022generating, ao2022rhythmic}, we perform one-hot encoding of the speakers' ID as $F^\texttt{I} \in \{0, 1\}^{N_I}$, where $N_I$ represents the number of speakers in the dataset.

\noindent\textbf{Audio Encoding: }
We first encode the input audio clip $A$ with MFCC~\cite{sahidullah2012design}, yielding an initial set of per-frame features $F^M = [F^M_1, F^M_2, \cdots, F^M_T]\in \mathbb{R}^{80\times T}$.

In order to decouple semantic components of $A$, we adapt the pre-trained model of StyleTTS~\cite{li2022styletts}, which also takes hand-crafted features from MFCC as input.
As a by-product, the features extracted by StyleTTS naturally consist of three components, yielding
\[F^A =  [F^A_1, F^A_2, \cdots, F^A_T]\in\mathbb{R}^{130\times T}, \mbox{where } F^A_i = [e_i; p_i; y_i], \]
where $e_i, p_i\in \mathbb{R}^1, y_i\in \mathbb{R}^{128}$ correspond the features regarding energy, pitch, and style of the $i-$frame of $A$, respectively. 
The former two reflect the intensity and pitch of the sound, while the latter encodes the emotional aspects of the audio. 
The extracted features have a strong correlation with the body and hand motions. 
As will be shown in Sec.~\ref{sec:speaker}, such feature decomposition facilitates fine-grained control over the generated body language. 

\noindent\textbf{Text Encoding: } Transcription is a crucial form of speech representation that conveys detailed linguistic information in a concise format. It is usually presented as a sequence of words, and the rate at which words are spoken can vary with the speech tempo. To address this variability, we follow~\cite{ao2022rhythmic} to align words with the corresponding speech and convert the text into frame-level features. Subsequently, we take the aligned text as input and use the pre-trained model of BERT~\cite{devlin2018bert} to encode $W$, obtaining $F^W = [F_1^W, F_2^W, \cdots, F_T^W]\in \mathbb{R}^{768\times T}$. We obtain the aligned phoneme in a similar manner which reflects the content information of the audio and input it into the text encoder of StyleTTS to obtain the phoneme feature $F^P = [F_1^P, F_2^P, \cdots, F_T^P]\in \mathbb{R}^{512\times T}$.

In the end, we denote by $F^\texttt{m}$ the extracted features from the input verbal cues $A$ and $W$, and $F^\texttt{f}$ from $A$ and $P$: 
\begin{equation}\label{eqn:feat_alt}
\begin{split}
    F^\texttt{m} = [F^A; F^W] \in \mathbb{R}^{898 \times T}, \\
    F^\texttt{f} = [F^A; F^P] \in \mathbb{R}^{642 \times T}.
\end{split}
\end{equation}



\subsection{VQ-VAE Model}\label{sec:vq}
We first denote the SMPL-X p-GT label for the speaker and listener over a time duration $T$ by $S_{1:T}=[s_1, ..., s_T]$ and $L_{1:T}=[l_1, ..., l_T]$, respectively. 
For each $s_i$ ($l_j$), we indicate the label regarding facial expressions, body movements, and hand gestures by superscripts $\f$, $\bo$, and $\h$, respectively. 
That is $s_i = [s_i^{\f}; s_i^{\bo}; s_i^{\h}]$ (similar definition for $l_j$). 
For the sake of simplicity, we also stack the label along each attribute, for example, $S^{\f} = [s_1^{\f}, s_2^{\f}, \cdots, s_T^{\f}]$.


Given a raw audio clip $A$, we first obtain $F^\texttt{m}$ and $F^{\texttt{f}}$ via Eqn.(~\ref{eqn:feat_alt}). The speaker's lip shape and expression are generally related to the phonemes, rhythm, and other information in the audio~\cite{wu2023speech2lip}. For the generation of facial expressions for the speaker, we follow~\cite{yi2022generating} and model it as a regression task. 
More specifically, we train a network, $\mathcal{N}_{S}^{\f}$, to recover $\hat{S^{\f}}$, namely
\begin{equation}\label{equ:face}
    \hat{S^{\f}} = \mathcal{N}_{S}^{\f}(F^{\texttt{f}}, F^{\texttt{I}})=[\hat{s_1}^{\f}, \hat{s_2}^{\f}, \cdots, \hat{s_T}^{\f}].   
\end{equation}
We use CNN~\cite{yi2022generating} as the backbone of $\mathcal{N}_{S}$ and Mean Squared Error (MSE) loss for training.  

On the other hand, we aim to generate body and hand motions for speakers and full-body movements for listeners based on verbal input. 
Apart from maximizing diversity as in general generation tasks, we further emphasize the harmony between the generated body and hand motions. 
To this end, we utilize a VQ-VAE~\cite{van2017neural} network and employ an autoencoder to discretely encode the regarding attributes. We take the generation of the speaker's body motion as an example. Given a sequence $S^{\bo}= [s_1^{\bo}, s_2^{\bo}, \cdots, s_T^{\bo}]$, we train an autoencoder and a finite-dimensional codebook $C_{S}^{\bo}=[c_{S, 1}^{b},...,c_{S, K}^{b}]$, which contains $K$ codes. 

We denote the encoder and decoder of the VQ-VAE as $\mathcal{E}^{\bo}_S$ and $\mathcal{D}^{\bo}_S$ respectively. 
The former takes as input $S^{\bo}$, and output latent code $Z_{S}^{\bo}=[z_{S, 1}^{\bo},...,z_{S, \gamma}^{\bo}]$, where $\gamma = T/w$ and $w$ is the temporal window size. 
Then we can quantize the $i$-th embedding $z_{S, i}^{b}$ by comparing it with the codebook, $C_{S}^{\bo}$, to find the closest code:
\begin{equation}\label{eqn:enc}
    \hat{z}_{S, i}^{\bo} = \mathop{\arg\min}\limits_{c_{S, k}^{\bo} \in C_{S}^{\bo}}\Vert z_{S, i}^{\bo} - c_{S, k}^{\bo}\Vert.
\end{equation}
Then, we reconstruct the body motion with the decoder, with respect to the latent code in Eqn.(\ref{eqn:enc}): $\hat{S}^{\bo} = \mathcal{D}_{S}^{\bo}(\hat{Z}_{S}^{\bo})$. 

Finally, we train the autoencoder and the codebook using the following loss function~\cite{van2017neural}:
\begin{equation}
\begin{split}
    L_{\mbox{vq}} = L_{\mbox{rec}}(\hat{S}^{\bo}, S^{\bo}) + \Vert sg[Z_{S}^{\bo}] - \hat{Z}_{S}^{\bo}\Vert \\
    + \beta\Vert Z_{S}^{\bo} - sg[\hat{Z}_{S}^{\bo}]\Vert, 
\end{split}
\end{equation}
where $L_{\mbox{rec}}$ denotes the MSE loss, $sg\left[\cdot\right]$ is the stop-gradient operator and $\beta$ is a hyper-parameter for the commitment loss. 
These components ensure that the encoder commits to specific codes and that the codebook is utilized optimally. 
We independently train the autoencoder and the codebook for each of $S^{\h}, L^{\f}, L^{\bo}, L^{\h}$, which are denoted in the same manner as above in the following. 

\subsection{Architecture of Speaker-Listener Generator}\label{sec:autoreg}
With the learned VQ-VAEs in Sec~\ref{sec:vq}, the motion of the speaker $S_{1:T}$ and listener $L_{1:T}$ can be encoded as a sequence of indices $I_{S, 1:\gamma}$ and $I_{L, 1:\gamma}$ obtained via Eqn.(\ref{eqn:enc}), with respect to the corresponding encoder and codebook.  
We can project these indices back to their corresponding codebook entries and decode the obtained latent code to reconstruct the motion. 
Therefore, we model motion generation as an auto-regressive next-index prediction from audio input. 
In particular, we not only consider the coordination between the facial expressions and body movements of the speaker and the listener individually but also account for the mutual influence between them. 
To this end, we need to exploit the interaction between the speaker and the listener. 
More specifically, we formulate the generation of the speaker's and listener's motion as
\begin{eqnarray}\label{equ:reg}
    p(I_{S}|F^\texttt{m}, F^\texttt{I}) = \prod \limits_{i=1}^{|I_{S}|}p(I_{S, i}|F^\texttt{m}, F^\texttt{I}, I_{S, 1:i-1}, I_{L, 1:i-1}), \\
    p(I_{L}|F^\texttt{m}, F^\texttt{I}) = \prod \limits_{i=1}^{|I_{L}|}p(I_{L, i}|F^\texttt{m}, F^\texttt{I}, I_{S, 1:i}, I_{L, 1:i-1}), 
\end{eqnarray}
where $I$ is assembled by $I^{\bo}, I^{\h}, I^{\f}$, 
representing the motion parameters of the body, hand, and face for either the speaker or the listener. 

We highlight the difference between our chain-like design and that of~\cite{yi2022generating}. 
The latter only considers the generation task for a speaker, which solely enforces the harmony among the generation of the speaker's body and hand. 
On the other hand, our design further
allows for the simultaneous output of both the speaker and listener and takes into consideration the mutual influence between them while the conversation is going on. 

We train an auto-regressor with a transformer backbone~\cite{zhang2023t2m} in this generation network, where we maximize the log-likelihood of the data distribution:
\begin{equation}
    L_{\mbox{reg}} = E_{{I \sim p(I)}}[{-}\log p(I|F^\texttt{m}, F^\texttt{I})].
\end{equation}


\section{Dataset Construction}
In this section, we introduce our new dataset, named \texttt{HoCo}, which is designed for holistic non-verbal communication. The \texttt{HoCo} dataset comprises video clips sourced from natural settings. It encompasses both verbal elements (audio clips and their transcripts) and 3D annotations of non-verbal signals like facial expressions, body postures, and hand gestures. Distinctively, our dataset provides data for both speakers and listeners, unlike previous works that focus on one or the other. Additionally, \texttt{HoCo} includes annotations of listeners' emotions. Although not used in our experiments, this feature adds more depth compared to other leading datasets such as SHOW~\cite{yi2022generating}.

\begin{figure}[t!]
  \begin{center}
\includegraphics[width=8.5cm]{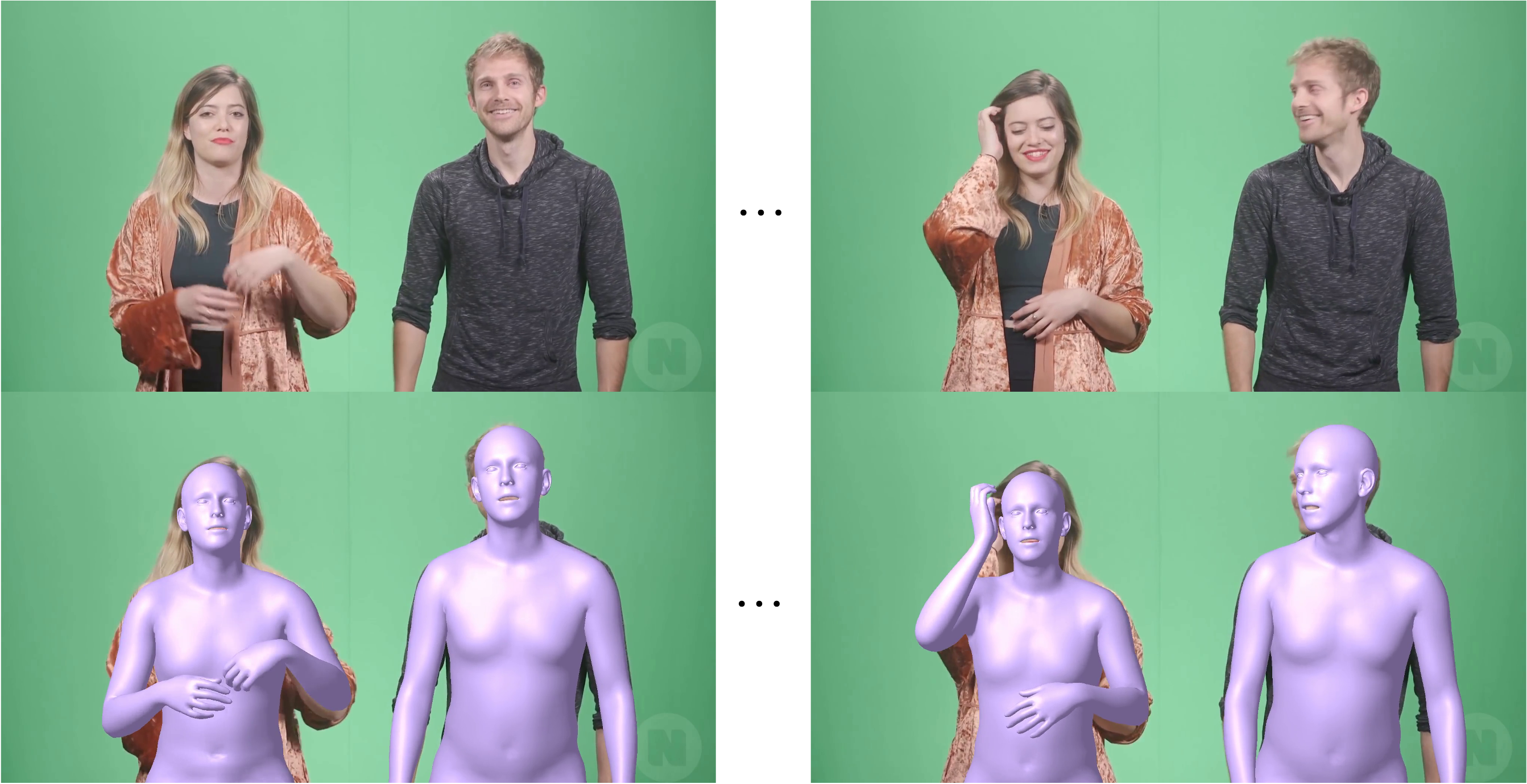}
  \end{center}
\caption{In \texttt{HOCO}, we provide high-definition videos of two-person communication (top), as well as the corresponding p-GT estimated by SMPL-X (bottom).}
\label{fig:dataset_show}
\end{figure}

Tab.~\ref{table:data} presents a clear comparison between the existing benchmarks and ours. 
We emphasize that our dataset enjoys several features including being captured in a real-world scenario, carrying rich annotations, and involving the most comprehensive information on speaker-listener interaction among all. In Fig.~\ref{fig:dataset_show}, we select one scene from our dataset to showcase, which include the RGB images of the videos and the corresponding visualizations of SMPL-X reconstructions. In the following, we provide details on our dataset construction. 

\subsection{Raw Data Collection and Screening}
We start by collecting conversational video clips from YouTube\footnote{https://www.youtube.com} involving two participants. 
The scenes include dual hosts, two-person blog programs, and talk show programs. 
We have curated a total of $45$ hours of video data, which includes $22,913$ clips involving $26$ speaker IDs.
The selected clips pass our screen process based on the following criteria: 1) The video frames should feature two individuals, of which one is speaking, and the other responds accordingly; 
2) Both individuals' upper bodies are clear and unobstructed, with arms fully visible during their communication, and their facial expressions are clear; 
3) There is clear and noticeable interaction between the participants. The clips with listeners showing significant reactions such as nodding, smiling, shrugging, and rich facial expressions such as happiness or surprise are preferred. 

\subsection{Annotation Procedure}

Unlike existing datasets such as SHOW~\cite{yi2022generating}, we provide multi-modal data including audio clips, aligned transcripts, and the pseudo-GroundTruth (p-GT) labels with respect to SMPL-X~\cite{SMPL-X:2019} model. 

\noindent\textbf{Audio and Transcript Processing:} Based on high-resolution RGB videos, we initially extract audio using the $\textbf{{moviepy}}$\footnote{https://pypi.org/project/moviepy/} library in Python. Subsequently, based on the audio, we use the $\textbf{{whisper}}$\footnote{https://github.com/openai/whisper} library to extract the corresponding transcripts. To align each transcript with the corresponding time frames, we employ Montreal Forced Aligner (MFA)~\cite{mcauliffe2017montreal} to detect the start and end times of each word. 

\noindent\textbf{3D Body Language Annotation: }
Our dataset also provides 3D whole-body meshes as pseudo-labels. 
We choose SMPL-X~\cite{SMPL-X:2019} as the p-GT for the dataset because it contains rich facial, hand, and body details. 
We use PyMFA-X~\cite{pymafx2023} 
which is efficient and accurate for extracting SMPL-X parameters from a video clip. 
To ensure the accuracy of GT generation, we first use Detectron2~\cite{wu2019detectron2} to separately detect the masks for the speaker and listener, and then generate the corresponding SMPL-X p-GTs. The p-GT comprises parameters of a shared body shape $\beta \in \mathbb{R}^{10}$, poses $\theta_h \in \mathbb{R}^{156 \times T}$, a shared camera pose $\theta_c \in \mathbb{R}^{3}$ and translation $\epsilon \in \mathbb{R}^{3}$, and facial expressions $\phi \in \mathbb{R}^{50 \times T}$.
\section{Experiments}

\subsection{Datasets}
We start by detailing the datasets and evaluation metrics used in our experiments. 
Since our goal is to generate full body language, apart from our proposed dataset, \texttt{HoCo}, we consider the SHOW dataset for benchmarking our design on the single-role pipeline: 

\noindent\textbf{SHOW~\cite{yi2022generating}: } This dataset is a filtered version based on~\cite{habibie2021learning}, resulting $26.9$ hours of high-quality videos. The dataset comprises $4$ speaker IDs.
Additionally, p-GT for SMPL-X parameters is annotated. On this basis, we generate text aligned with the number of frames and filter out data that cannot be automatically aligned.

\noindent\textbf{HoCo: } We evaluate our method and the baselines with our \texttt{HoCo} dataset. More precisely, $\data$ dataset is built on video clips collected \emph{in the wild}, which provides holistic information of talks in the following two perspectives: 1) It includes both verbal inputs-- audio clips and accompanied transcripts) and 3D annotations for the non-verbal signals (facial expressions, body poses, and hand gestures); 2) Unlike the prior works biased towards either speakers or listeners, the input and annotation above are constructed for \emph{both} speakers and listeners within the collected talks. 
Moreover, we also provide annotation for listeners' emotions. 
Though this attribute is not utilized in our experiments, we highlight that $\data$ contains more modalities than the state-of-the-art datasets (\emph{e.g.}, SHOW~\cite{yi2022generating}). 


\subsection{Implementation Details}
The training is composed of two stages. For the VQ-VAE model, the codebook size is set to $256\times256$. The temporal window size $w$ is $4$ and we set the number of codes to be $2,048$. The weight, $\beta$, of the commitment loss term in Equation(4) in the main submission is set to be $0.25$. For the Speaker-Listener Generator, we choose a transformer model comprising $9$ layers and $16$ heads. We set the block size to be $376$. We adapt Adam with $\beta_1 = 0.9$, $\beta_2 = 0.999$, and a learning rate of $0.0001$ as the optimizer.
For all experiments, the batch size is set to $128$, and training is conducted on a single Tesla V100-32G GPU for $100$ epochs with all video clips cropped into $88$ frames.

\subsection{Evaluation Metrics}
We roughly categorize the involved evaluation metrics into deterministic ones and non-deterministic ones. 

The former is especially applied to the task of generating the speaker's facial expressions, which is performed as a deterministic task. 
In particular, we use $\mathbf{L2}$ and Landmark Velocity Difference (\textbf{LVD}) to measure the authenticity and accuracy of generated expressions. 
The $\mathbf{L2}$ distance is to measure the difference between the generated facial expression and the ground truth label. 
\textbf{LVD} quantifies velocities of corresponding facial expressions between the ground truth and the generated expressions. 
This measurement helps to evaluate how well a model captures the dynamic aspects of facial movements.

Regarding the latter, we follow~\cite{zhi2023livelyspeaker} to assess the body and hand motions of both speaker and listener. 
The evaluation metrics include Frechet Gesture Distance (\textbf{FGD})~\cite{yoon2020speech}, Beat Consistency Score (\textbf{BC})~\cite{liu2022learning}, \textbf{Variation}~\cite{yi2022generating}, Concordance Correlation Coefficient (\textbf{CCC})~\cite{song2023react2023}, and Time Lagged Cross Correlation (\textbf{TLCC})~\cite{boker2002windowed, ng2022learning}. \textbf{FGD} provides a quantitative measure of the dissimilarity between the generated and ground truth motions which uses a pre-trained autoencoder to project the motion into latent space. \textbf{BC} is used to assess the consistency of motion, particularly in the context of body and hand movements associated with rhythmic beats or gestures. 
We utilize both \textbf{FGD} and \textbf{BC} together to evaluate the rationality of the generated motion. \textbf{Variation} is calculated by the variance across the time series sequence of the motions.

Meanwhile, there is no golden standard for evaluating whether a listener's reaction is appropriate in response to a speaker's behavior, while manually labeling would be labor-intensive and subjective. Therefore, within the evaluation metrics for the generation of the listener's full-body motions, we incorporate \textbf{CCC} followed REACT23 challenge~\cite{song2023react2023}, to calculate the correlation between the generated full-body motions and its most similar appropriate motion. Additionally, we also employ \textbf{TLCC} to measure the synchrony between the generated speaker and listener for dyadic motion assessment.

\subsection{Baselines}
Regarding generating speakers' motion from speech, 
we compared our method with the SOTA methods \textbf{Habibie et al.}~\cite{habibie2021learning} and \textbf{TalkSHOW}~\cite{yi2022generating} in the speech-to-motion task. 
Simultaneously, we construct some baselines to validate the effectiveness of our approach. 
In the following, \textbf{Audio + CVAE} refers to the network structure of~\cite{petrovich21actor}, using audio as a condition to generate corresponding motion through a VAE autoencoder structure. \textbf{TalkSHOW + ours feature} replaces the audio processing method of TalkSHOW with the decoupled feature approach we proposed. 

For our novel task of non-verbal communication generation with the \texttt{HoCo} dataset, as there exists no available baseline, we follow the responsive listening head generation task proposed in~\cite{zhou2022responsive}, which uses the audio and speaker's information as conditions for generating the listener. We compare with our formulated variants instead. Namely, for a baseline $\mathbf{X}$, we first train the model to generate the speaker with $\mathbf{X}$ and use the SMPL-X p-GT corresponding to the speaker as the condition to generate the listener. We refer to the results of this approach as $\mathbf{X \mbox{-split}}$.

\begin{figure}[b!]
  \begin{center}
\includegraphics[width=8.5cm]{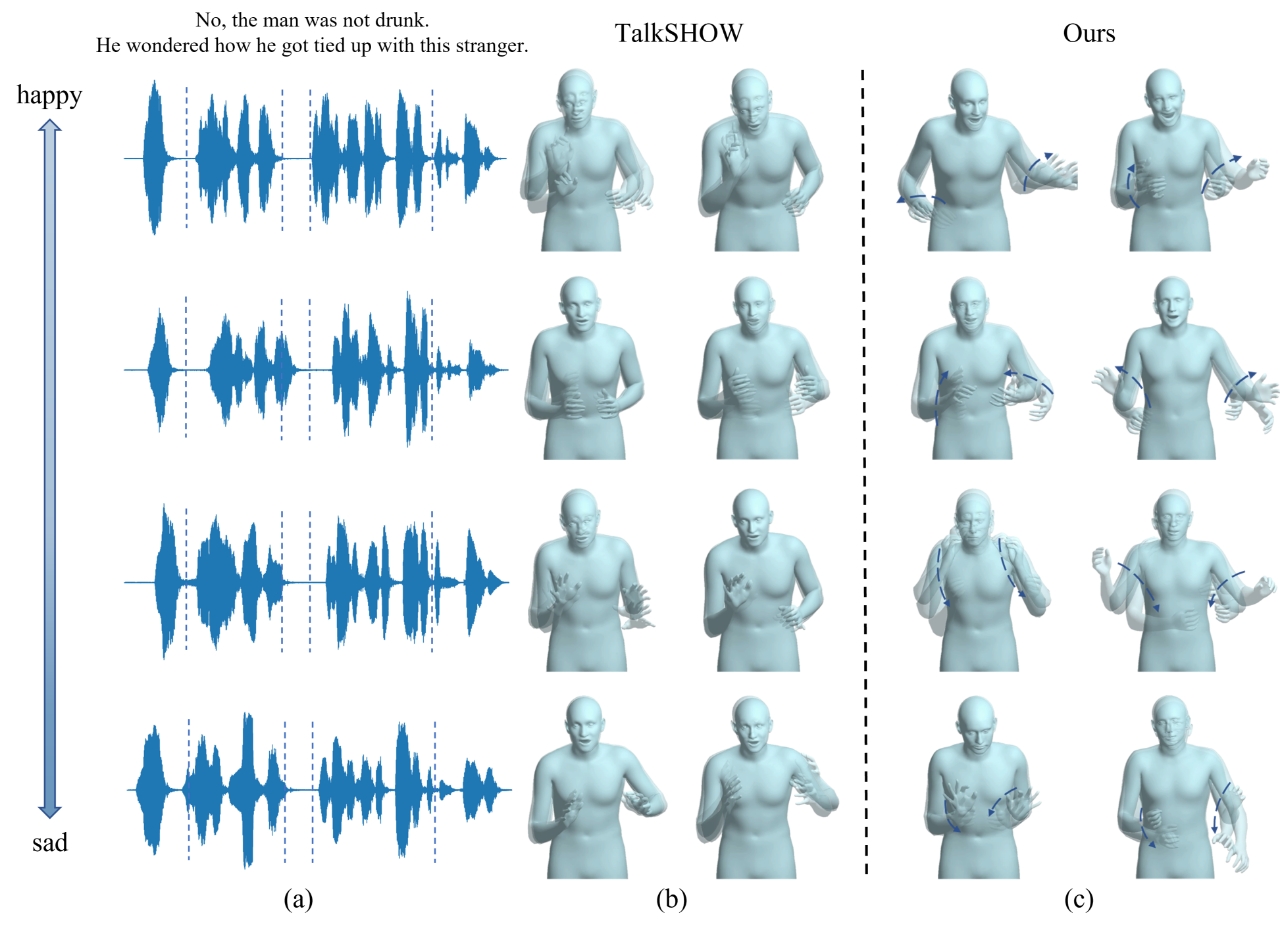}
  \end{center}
\caption{(a) shows a piece of transcript and the corresponding audio signals with varying pitches and emotions. (b) displays the inference results from TalkSHOW~\cite{yi2022generating}, where the generated motions have low sensitivity to changes in the audio signals. (c) presents our inference results, demonstrating high consistency between the generated motions and the audio.}
\label{fig:decouple}
\end{figure}

\subsection{Decoupling Feature Results}
In Section 3.2, we introduce an audio decoupling method, and now we provide a more detailed qualitative validation of its effects. If the input contains only textual information, the corresponding generated results should be similar. However, if the same sentence is spoken by a speaker with different pitches or emotions, there should be distinct motion expressions. Based on this, we select the dataset provided by~\cite{kaisiyuan2020mead}, where the same text includes two speaking emotions, \textbf{happy} and \textbf{sad}, each with two intensity levels from low to high. We use the model trained on the SHOW dataset~\cite{yi2022generating} to directly infer these unseen audios. We compare our method with TalkSHOW~\cite{yi2022generating}, and the results are shown in Fig.~\ref{fig:decouple}.

Our method, built upon the foundation of textual semantic information, explicitly considers audio pitch and style elements, where style represents the emotional content of the audio. Our results show significant variation in gestures generated for different audio emotions. For instance, gestures associated with happiness are predominantly upward, while those for sadness tend toward downward movements. With the same emotion at different pitches, a higher pitch variation corresponds to a more pronounced motion. In contrast, results generated by TalkSHOW~\cite{yi2022generating} for the same text under different audio conditions exhibit minimal variation in motion, indicating poorer consistency with the audio.

\subsection{Speaker-Centric Generation}
\noindent\textbf{Quantitative Analysis: }\label{sec:speaker}
Tab.~\ref{table:1} shows the quantitative results on the SHOW dataset. We first validate the effectiveness of our proposed feature decoupling. Using the model framework from TalkSHOW~\cite{yi2022generating} as a baseline, we replace the input features with our decoupled features. 
In the generation of facial expressions, compared to directly extracting features from a pre-trained wav2vec~\cite{baevski2020wav2vec} model, our decoupled features result in more accurate expressions. 
Moreover, our features assist in generating body and hand movements that are more similar to the ground truth. 
In comparison to baselines, our generated results exhibit an improvement both in audio consistency and motion diversity.
\begin{table}[!b]
\caption{Comparison to baselines on SHOW dataset~\cite{yi2022generating}. ↑ indicates higher is better and ↓ indicates lower is better.}\label{table:1}
\centering
\begin{tabular}{lccc}
\hline
\rowcolor[HTML]{FFFFFF} 
\multicolumn{1}{c}{\cellcolor[HTML]{FFFFFF}}                                  & \multicolumn{3}{c}{\cellcolor[HTML]{FFFFFF}Face} \\
\rowcolor[HTML]{FFFFFF} 
\multicolumn{1}{c}{\multirow{-2}{*}{\cellcolor[HTML]{FFFFFF}Method}} & L2 ↓                    &                & LVD ↓          \\ \hline
\rowcolor[HTML]{FFFFFF} 
Habibie et al.~\cite{habibie2021learning}                                                                & 0.237                        &                & 0.036               \\
\rowcolor[HTML]{FFFFFF} 
TalkSHOW~\cite{yi2022generating}                                                                      & 0.215                   &                & 0.032          \\
\rowcolor[HTML]{E7E6E6} 
Ours                                                                          & \textbf{0.209}          & \textbf{}      & \textbf{0.031} \\ \hline
\rowcolor[HTML]{FFFFFF} 
& \multicolumn{3}{c}{\cellcolor[HTML]{FFFFFF}Body\&Hands}   \\
\rowcolor[HTML]{FFFFFF} 
& FGD ↓                   & BC ↑           & Variation ↑    \\ \hline
\rowcolor[HTML]{FFFFFF} 
Habibie et al.~\cite{habibie2021learning}                                                                & 3.198                        &  0.452              & 0.051               \\
\rowcolor[HTML]{FFFFFF} 
Audio+CVAE~\cite{petrovich2021action}                                                                    & 2.561                   & 0.665          & 0.134          \\
\rowcolor[HTML]{FFFFFF} 
TalkSHOW~\cite{yi2022generating}                                                                      & 2.049                   & 0.847          & 0.332          \\
\rowcolor[HTML]{FFFFFF} 
TalkSHOW+our features                                                         & 1.902 &  0.849      &    0.328            \\
\rowcolor[HTML]{E7E6E6} 
\textbf{Ours}                                                                 & \textbf{1.483}          & \textbf{0.854} & \textbf{0.617} \\ \hline
\end{tabular}
\end{table}

\begin{figure}[t!]
  \begin{center}
\includegraphics[width=8.5cm]{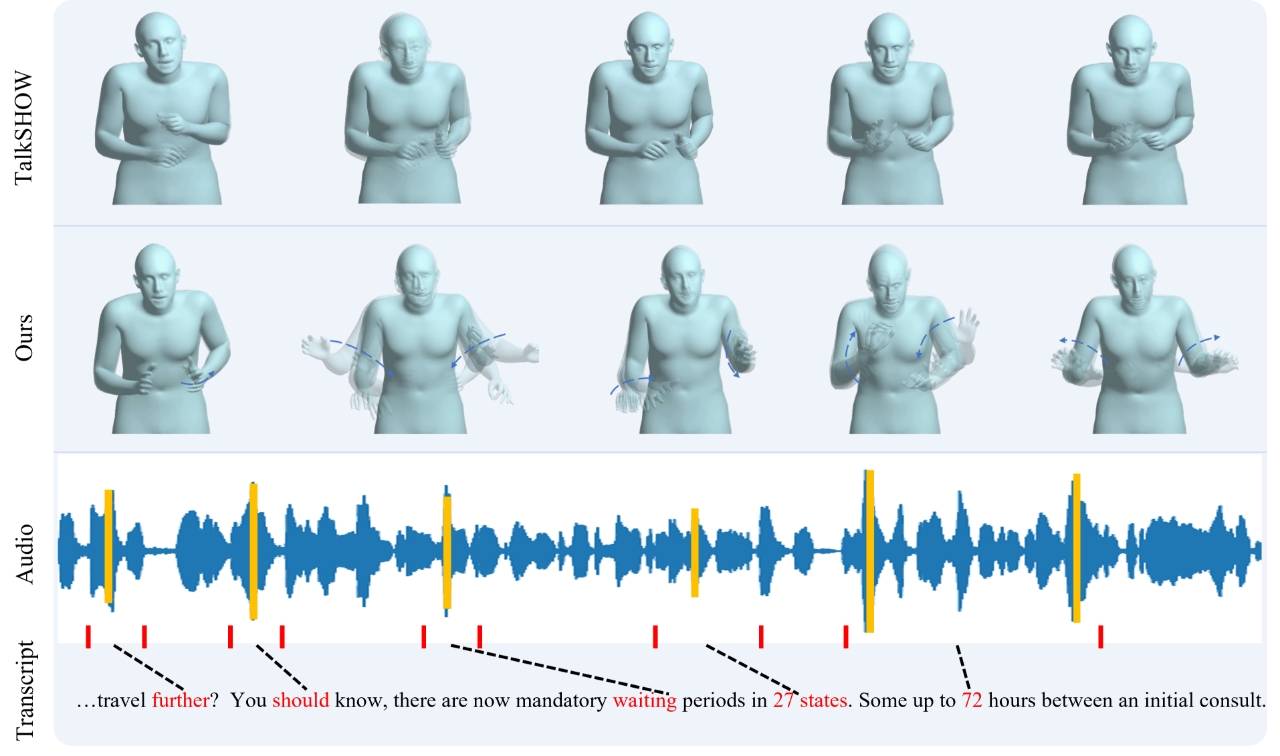}
  \end{center}
\caption{Visual comparisons with the TalkSHOW~\cite{yi2022generating} baseline on SHOW dataset. Our method generates diverse motions consistent with the rhythm of the input audio. In this audio clip with progressively emotional content, our generated results exhibit more diverse motions.}
\label{fig:single}
\end{figure}

\noindent\textbf{Qualitative Analysis: }
Fig.~\ref{fig:single} shows a comparison between the motion generated by us and those of TalkSHOW~\cite{yi2022generating} for a speech clip with progressive emotions. 
The sentence contains some emphasized words, including ``should" ``waiting" etc., accompanied by intonation changes. 
Our generated results exhibit more noticeable body movement changes in these areas. 
For emotionally progressive words like ``27" and ``72", our results introduce gestures such as shrugging. 
In contrast, results generated by TalkSHOW~\cite{yi2022generating} lack diversity, with only small, regular hand movements. 
Our results are more vivid in comparison. We also provide a demo video as a supplement.

\subsection{Speaker and listener Generation}
\noindent\textbf{Quantitative Analysis: }
For the speaker and listener generation framework, we specifically design baselines to validate the effectiveness of our proposed framework. 
The quantitative results are shown in Tab.~\ref{table:2}. 
Firstly, we observe the improvement of \textbf{TalkSHOW + our features\mbox{-split}} upon \textbf{TalkSHOW\mbox{-split}}, which confirms again the advantage of our factorized feature. 
It is also consistent with the results reported in Tab.~\ref{table:1}. 
Secondly, the performance gap between \textbf{Ours\mbox{-split}} with \textbf{Ours} validates the effectiveness of our chain-design auto-regressor in improving the diversity of generated results. 
Finally, compared to our proposed variants on top of external baselines, including \textbf{TalkSHOW\mbox{-split}}, \textbf{TalkSHOW + our features\mbox{-split}} and \textbf{Ours\mbox{-split}}. 
Ours achieves a 2.4\% improvement in FGD for the speaker and a 24.9\% for the listener. The diversity of our results has also seen improvement. In the generation of listener actions, we consider CCC to assess the appropriateness of the generated motions. Our method achieves a 13.4\% improvement, which indicates that our approach can generate not only diverse listener's motions but also actions that are more in line with reality. At the same time, taking into account the structure of both the speaker and the listener results in an 18.1\% increase in synchrony within the generated outcomes.

\begin{figure}[!b]
  \begin{center}
\includegraphics[width=8.5cm]{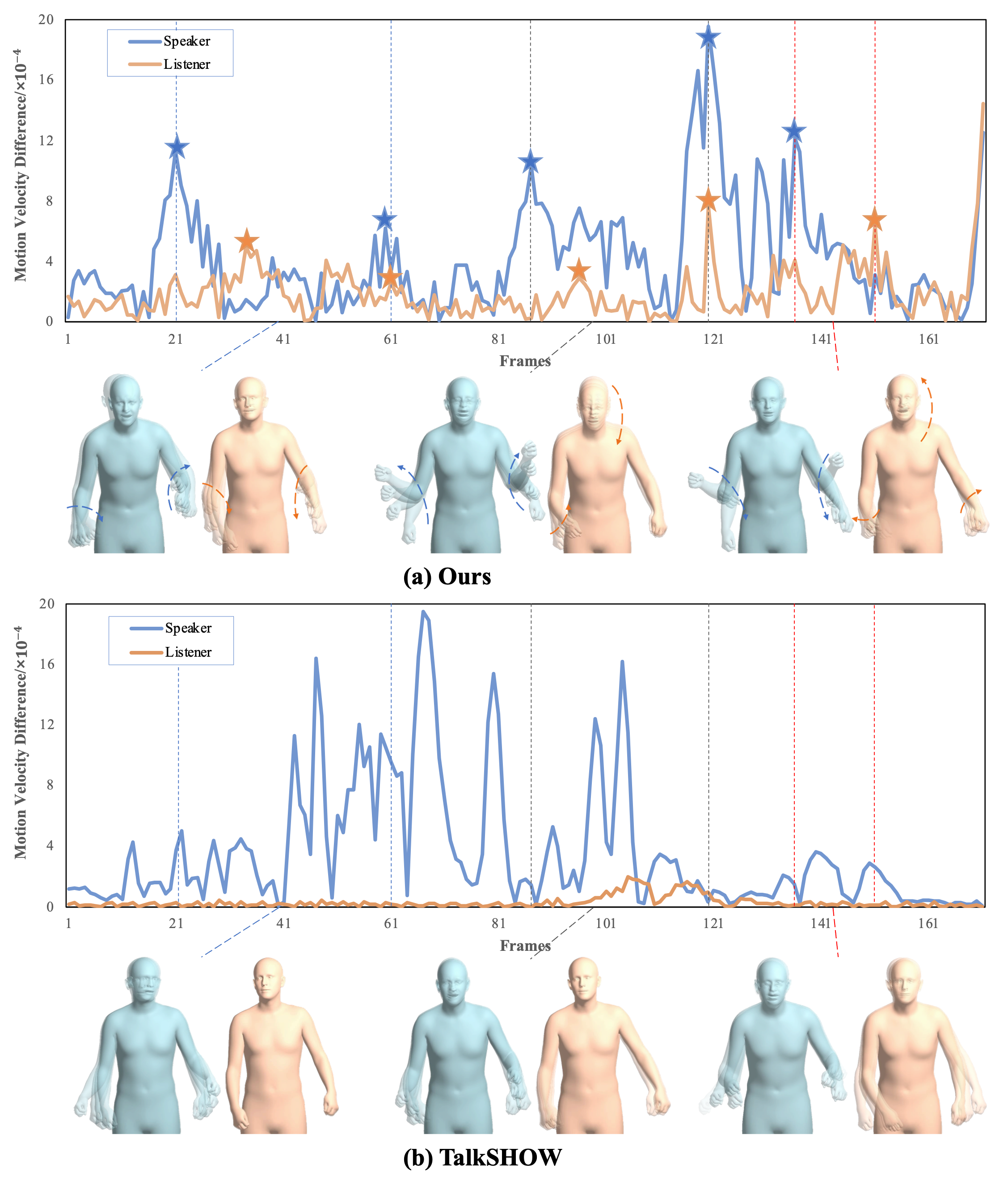}
  \end{center}
\caption{Visual comparisons with the TalkSHOW baseline on \texttt{HoCo} dataset. Our approach can generate semantically meaningful movements such as \textbf{laughing} and \textbf{stooping}.}
\label{fig:double}
\end{figure}

\begin{table*}[!t]
\caption{Comparison to baselines on \texttt{HoCo} dataset. ↑ indicates higher is better and ↓ indicates lower is better.}\label{table:2}
\centering
\setlength{\tabcolsep}{16pt}
\resizebox{\textwidth}{13mm}{
\begin{tabular}{lccc|cccc|c}
\hline
\rowcolor[HTML]{FFFFFF} 
\multicolumn{1}{c}{\cellcolor[HTML]{FFFFFF}}                         & \multicolumn{3}{c|}{\cellcolor[HTML]{FFFFFF}Speaker} & \multicolumn{4}{c|}{\cellcolor[HTML]{FFFFFF}Listener}             & Speaker\&Listener \\
\rowcolor[HTML]{FFFFFF} 
\multicolumn{1}{c}{\multirow{-2}{*}{\cellcolor[HTML]{FFFFFF}Method}} & FGD ↓            & BC ↑            & Variation ↑     & FGD ↓          & BC ↑           & Variation ↑    & CCC ↑           & TLCC ↓             \\ \hline
\rowcolor[HTML]{FFFFFF} 
Habibie et al. split                                                 & 0.835            & 0.511           & 0.022           & 0.674          & 0.566          & 0.010           & 0.082          & 28.740             \\
\rowcolor[HTML]{FFFFFF} 
Audio+CVAE split                                                     & 0.684            & 0.720            & 0.063           & 0.394          & 0.656          & 0.030           & 0.090           & 25.316            \\
\rowcolor[HTML]{FFFFFF} 
TalkSHOW split                                                       & 0.593            & 0.834           & 0.333           & 0.435          & 0.887          & 0.126          & 0.118          & 21.947            \\
\rowcolor[HTML]{FFFFFF} 
TalkSHOW+our features   split                                        & 0.572            & 0.833           & 0.414           & 0.342          & 0.871          & 0.135          & 0.121          & 21.632            \\
\rowcolor[HTML]{FFFFFF} 
Ours split                                                           & 0.761            & 0.901           & 0.469           & 0.337          & \textbf{0.898} & 0.210           & 0.162          & 19.628            \\
\rowcolor[HTML]{E7E6E6} 
\textbf{Ours}                                                        & \textbf{0.558}   & \textbf{0.906}  & \textbf{0.492}  & \textbf{0.253} & 0.883          & \textbf{0.261} & \textbf{0.187} & \textbf{16.076}   \\ \hline
\end{tabular}}
\vspace{-0.5em}
\end{table*}

\begin{table*}[!t]
\caption{Ablation study on \texttt{HoCo} dataset. ↑ indicates higher is better and ↓ indicates lower is better.}\label{table:abl}
\centering
\setlength{\tabcolsep}{16pt}
\resizebox{\textwidth}{12mm}{
\begin{tabular}{lccc|cccc|c}
\hline
\rowcolor[HTML]{FFFFFF} 
\multicolumn{1}{c}{\cellcolor[HTML]{FFFFFF}}                           & \multicolumn{3}{c|}{\cellcolor[HTML]{FFFFFF}Speaker} & \multicolumn{4}{c|}{\cellcolor[HTML]{FFFFFF}Listener}             & Speaker\&Listener \\
\rowcolor[HTML]{FFFFFF} 
\multicolumn{1}{c}{\multirow{-2}{*}{\cellcolor[HTML]{FFFFFF}Ablation}} & FGD ↓            & BC ↑            & Variation ↑     & FGD ↓          & BC ↑           & Variation ↑    & CCC ↑          & TLCC ↓            \\ \hline
\rowcolor[HTML]{FFFFFF} 
Ours w/o text feature                                                  & 0.569            & 0.906           & 0.514           & 0.354          & 0.876          & 0.259          & 0.164          & 18.032            \\
\rowcolor[HTML]{FFFFFF} 
Ours w/o style feature                                                 & 0.591            & 0.901           & \textbf{0.520}  & 0.305          & 0.881          & \textbf{0.276} & 0.156          & 17.623            \\ \hline
\rowcolor[HTML]{FFFFFF} 
Ours w/o   chain-design                                                & 0.864            & 0.889           & 0.474           & 0.421          & 0.879          & 0.225          & 0.142          & 20.265            \\
\rowcolor[HTML]{FFFFFF} 
Ours w/o s-l chain-design                                              & 0.761            & 0.901           & 0.469           & 0.337          & \textbf{0.898} & 0.210          & 0.162          & 19.628            \\ \hline
\rowcolor[HTML]{E7E6E6} 
\textbf{Ours}                                                          & \textbf{0.558}   & \textbf{0.906}  & 0.492           & \textbf{0.253} & 0.883          & 0.261          & \textbf{0.187} & \textbf{16.076}   \\ \hline
\end{tabular}}
\vspace{-0.5em}
\end{table*}

\noindent\textbf{Qualitative Analysis: }
We show an example from the speaker and listener generation experiment to demonstrate the coherence of our generated speaker and listener actions.
This is a playful scenario where, in the latter part of the audio, both the speaker and the listener are laughing. Fig.~\ref{fig:double}(a) above illustrates the variation in SMPL-X parameters generated by our approach for the speaker and listener, demonstrating a consistent change in parameters, as indicated by the stars. 
We also visualize corresponding significant changes for both in Fig.~\ref{fig:double}(a) below, where the listener responds with laughter and a bending posture following the speaker's movements, and the speaker, in turn, reacts with laughter. In contrast, Fig.~\ref{fig:double}(b) shows the corresponding results from TalkSHOW, where the listener's reactions are notably lacking.

\begin{table*}[!t]
\caption{The percentage of the user’s favorite methods in terms of natural, diversity, synchrony, and consistency.}\label{table:user}
\centering
\setlength{\tabcolsep}{16pt}
\resizebox{\textwidth}{8mm}{
\begin{tabular}{lccc|ccc|c}
\hline
\rowcolor[HTML]{FFFFFF} 
\multicolumn{1}{c}{\cellcolor[HTML]{FFFFFF}}                          & \multicolumn{3}{c|}{\cellcolor[HTML]{FFFFFF}Speaker}                                         & \multicolumn{3}{c|}{\cellcolor[HTML]{FFFFFF}Listener}                                        & \multicolumn{1}{l}{\cellcolor[HTML]{FFFFFF}Speaker\&Listener} \\
\rowcolor[HTML]{FFFFFF} 
\multicolumn{1}{c}{\multirow{-2}{*}{\cellcolor[HTML]{FFFFFF}Methods}} & Natural          & Diversity        & \multicolumn{1}{l|}{\cellcolor[HTML]{FFFFFF}Synchrony} & Natural          & Diversity        & \multicolumn{1}{l|}{\cellcolor[HTML]{FFFFFF}Synchrony} & Consistency                                                   \\ \hline
\rowcolor[HTML]{FFFFFF} 
TalkSHOW-split                                                        & 36.67\%          & 32.50\%          & 40.83\%                                                & 22.50\%          & 15.00\%          & 18.33\%                                                & 11.67\%                                                       \\
\rowcolor[HTML]{E7E6E6} 
\textbf{Ours}                                                         & \textbf{63.33\%} & \textbf{67.50\%} & \textbf{59.17\%}                                       & \textbf{77.50\%} & \textbf{85.00\%} & \textbf{81.67\%}                                       & \textbf{88.33\%}                                              \\ \hline
\end{tabular}
}
\vspace{-0.5em}
\end{table*}

\subsection{User Study}
We conduct a user study experiment to demonstrate the differences between our method and the SOTA method (TalkSHOW~\cite{yi2022generating}). We interview 12 volunteers and present them with 10 different cases of generated dual-person communication. The respondents evaluated the speaker and listener on three metrics: Natural, Diversity, and Synchrony, as well as the consistency of responses between the speaker and listener. We then calculate the percentage scores for each task across these metrics. As shown in Tab.~\ref{table:user}, our method is the most favored by participants in terms of the aforementioned criteria.

\subsection{Ablation study}
We conduct ablation experiments primarily focusing on the effects of decoupled features and different chaining designs in our proposed framework. The experiments are validated on the \texttt{HoCo} dataset, and the results are presented in Tab.~\ref{table:abl}. The first part of the table shows results when not using text and style-related features. The absence of the feature leads to a decrease in the coordination between generated motions and the ground truth. Simultaneously, using all features may result in a slight sacrifice in diversity. The second part validates the rationality of our chaining design. Initially, when we do not consider self-interactions or interactions between the speaker and the listener, 
as shown in \textbf{Ours w/o chain-design}, the FGD for the speaker increases 35.4\% and the FGD for the listener increases 39.9\%. Also, the CCC for the listener increases 24.1\%. On this basis, when we consider the interaction of their actions, as shown in \textbf{Ours w/o s-l chain-design}, compared to our design, there is an increase in FGD for both 26.7\% and 24.9\% and there is an increase in CCC for 13.4\%. At the same time, our design can enhance the diversity of both speaker and listener generations. The synchrony between the generated speaker and listener has a significant improvement.

\section{Conclusion}
In this paper, we take into consideration communication within human interaction and present a novel task that generates 3D holistic human motions for both speakers and listeners. 
To achieve this goal, we contribute to both dataset and model design. For the former, we provide the \texttt{HoCo} communication dataset for future exploration along this task.
For the latter, we propose a model tailored for our task, which consists of novel designs including 1) factorization for decoupling audio features which enhances the generation of more realistic and coordinated movements; 2) a chain-like auto-regressive model for characterizing non-verbal communication. Additionally, we achieve state-of-the-art performance on two benchmarks.

\noindent\textbf{Limitations} There is still room for improvement in our current research. Our data can only generate p-GT corresponding to the video data through algorithms. Also, we only consider the parametric human model of SMPL-X and do not provide the final realistic rendering results. Additionally, our work only considers situations where the positions of the speaker and the listener remain unchanged, and it cannot simulate scenarios where the positions of the speaker and listener change, which are also quite common in reality. In future work, we will consider a wider range of actions and verbal interactions between people, and we will look into combining different settings to generate realistic dyadic interactions.




{\small
\bibliographystyle{spbasic}
\bibliography{main}
}

\end{sloppypar}
\end{document}